\documentclass[letterpaper, 10 pt, conference]{ieeeconf}
\pdfoutput=1
\IEEEoverridecommandlockouts
\usepackage{amsmath}
\usepackage{hyperref} 
\usepackage{amssymb}
\usepackage{graphicx}
\usepackage{makecell}
\usepackage{mathptmx}
\usepackage{times}
\usepackage{xcolor}
\usepackage[capitalize,noabbrev]{cleveref}
\usepackage{caption}
\usepackage{xifthen}
\usepackage{xparse}
\usepackage{leftidx}
\usepackage{bm}
\usepackage{etoolbox}
\usepackage{multirow, makecell}
\usepackage{xspace}

\DeclareMathAlphabet{\mybf}{OT1}{ptm}{b}{n} 
\newcommand{\mybs}[1]{{\bm{#1}}} 
\DeclareMathAlphabet{\mybfi}{OML}{cmm}{b}{it}

\newcommand{\mbf}[1]{
\ifcat\noexpand#1\relax 
\mybs{#1}
\else
\mybf{#1}
\fi
}

\newcommand{\cframe}[1]{{\smash{\protect\underrightarrow{\mathcal{F}}_{#1}}}}

\newcommand{\vect}[1]{\mbf{\lowercase{#1}}}
\newcommand{\vecth}[1]{\hat{\mbf{\lowercase{#1}}}}

\newcommand{\N}{\mathbb{N}}
\newcommand{\R}{\mathbb{R}}
\newcommand{\SOthree}{SO(3)}

\newcommand{\true}{\textit{true}\xspace}
\newcommand{\seTwo}{\textit{supereight2}\xspace}
\newcommand{\SeTwo}{\textit{Supereight2}\xspace}
\newcommand{\glocal}{\textit{GLocal}\xspace}
\newcommand{\voxgraph}{\textit{Voxgraph}\xspace}

\newcommand{\free}{\textit{free}\xspace}
\newcommand{\occupied}{\textit{occupied}\xspace}
\newcommand{\unknown}{\textit{unknown}\xspace}
\newcommand{\frontiers}{\mathcal{F}}
\newcommand{\submaps}{\mathcal{M}}

\newcommand{\dimt}[1]{\textcolor{gray}{#1}}

\title{\LARGE \bf
Efficient Submap-based Autonomous MAV Exploration using Visual-Inertial SLAM Configurable for LiDARs or Depth Cameras
}

\author{Sotiris Papatheodorou$^{1,2,3,4,*}$, Simon Boche$^{1,*}$, Sebasti\'{a}n Barbas Laina$^{1,3}$, Stefan Leutenegger$^{1,2,3,4}$%
\thanks{This work was supported by the Technical University of Munich, MIRMI, the TUM Innovation Network CoConstruct, Leica Geosystems AG. and the EU Horizon projects DigiForest \scriptsize{(101070405)} and AUTOASSESS \scriptsize{(101120732)}.}%
\thanks{$^{1}$Smart Robotics Lab, School of Computation, Information and Technology, Technical University of Munich. E-mail addresses: \texttt{\{sotiris.papatheodorou, simon.boche, sebastian.barbas, stefan.leutenegger\}@tum.de}}%
\thanks{$^{2}$Smart Robotics Lab, Department of Computing, Imperial College London. E-mail addresses: \texttt{\{s.papatheodorou18, s.leutenegger\}@ic.ac.uk}}%
\thanks{$^{3}$Munich Institute of Robotics and Machine Intelligence (MIRMI).}%
\thanks{$^{4}$Munich Center for Machine Learning (MCML).}%
\thanks{$^{*}$ Equal contribution.}%
}

\begin{document}
\maketitle
\thispagestyle{empty}
\pagestyle{empty}

\begin{abstract}
Autonomous exploration of unknown space is an essential component for the deployment of mobile robots in the real world.
Safe navigation is crucial for all robotics applications and requires accurate and consistent maps of the robot's surroundings.
To achieve full autonomy and allow deployment in a wide variety of environments, the robot must rely on on-board state estimation which is prone to drift over time.
We propose a Micro Aerial Vehicle (MAV) exploration framework based on local submaps to allow retaining global consistency by applying loop-closure corrections to the relative submap poses.
To enable large-scale exploration we efficiently compute global, environment-wide frontiers from the local submap frontiers and use a sampling-based next-best-view exploration planner.
Our method seamlessly supports using either a LiDAR sensor or a depth camera, making it suitable for different kinds of MAV platforms.
We perform comparative evaluations in simulation against a state-of-the-art submap-based exploration framework to showcase the efficiency and reconstruction quality of our approach.
Finally, we demonstrate the applicability of our method to real-world MAVs, one equipped with a LiDAR and the other with a depth camera.
Video: \href{https://youtu.be/Uf5fwmYcuq4}{https://youtu.be/Uf5fwmYcuq4} 
\end{abstract}

\section{Introduction}
\label{chapter:introduction}
Autonomous exploration of unknown environments has been an active area of research in mobile robotics and remains a challenge.
It is essential for potential real-world robot applications, such as industrial or construction site inspection.
MAVs are a popular choice of platform due to their versatility, agility and ability to reach areas inaccessible to humans or ground robots.
The objective of exploration is commonly formulated as discovering and mapping an environment as fast as possible~\cite{yamauchi1997frontier,bircher2016receding,dai2020fast}. 
Ideally, to achieve full autonomy, the robot incrementally builds an accurate and complete map of its environment online, solely based on sensor inputs, that contains all information required for navigating through and localizing in it.
State-of-the-art approaches typically formulate this problem as identifying the next-best-pose that maximizes some utility and finding a collision-free path to the goal pose.

Simultaneous Localization and Mapping (SLAM) methods fusing complementary sensors are widely used for on-board pose estimation.
Visual-Inertial SLAM (VI-SLAM), combining visual and inertial measurements, offers high short-term accuracy but suffers from the accumulation of drift~\cite{OKVIS2,campos2021orb,usenko2019visual}.
Lately, driven by the availability of low-priced Light Detection and Ranging (LiDAR) sensors, LiDAR-Visual-Inertial SLAM (LVI-SLAM) has been a topic of extensive research~\cite{shan2021lvi,zheng2022fast,boche2024tightlycoupled} in the prospect of reducing drift.

In large-scale scenarios, the inevitable drift can lead to significant deterioration of the map accuracy.
One potential mitigation is using local submaps, based on the assumption that drift is negligible within a small region.
To maintain global consistency, the relative poses between individually rigid submaps can be updated upon loop closures~\cite{oxfordLidarSE} or incorporated into the state estimation problem~\cite{boche2024tightlycoupled,voxgraph}.

\begin{figure}[t]
    \centering
    \includegraphics[width=0.493\columnwidth,trim={7.5cm, 0, 11.1cm, 0},clip]{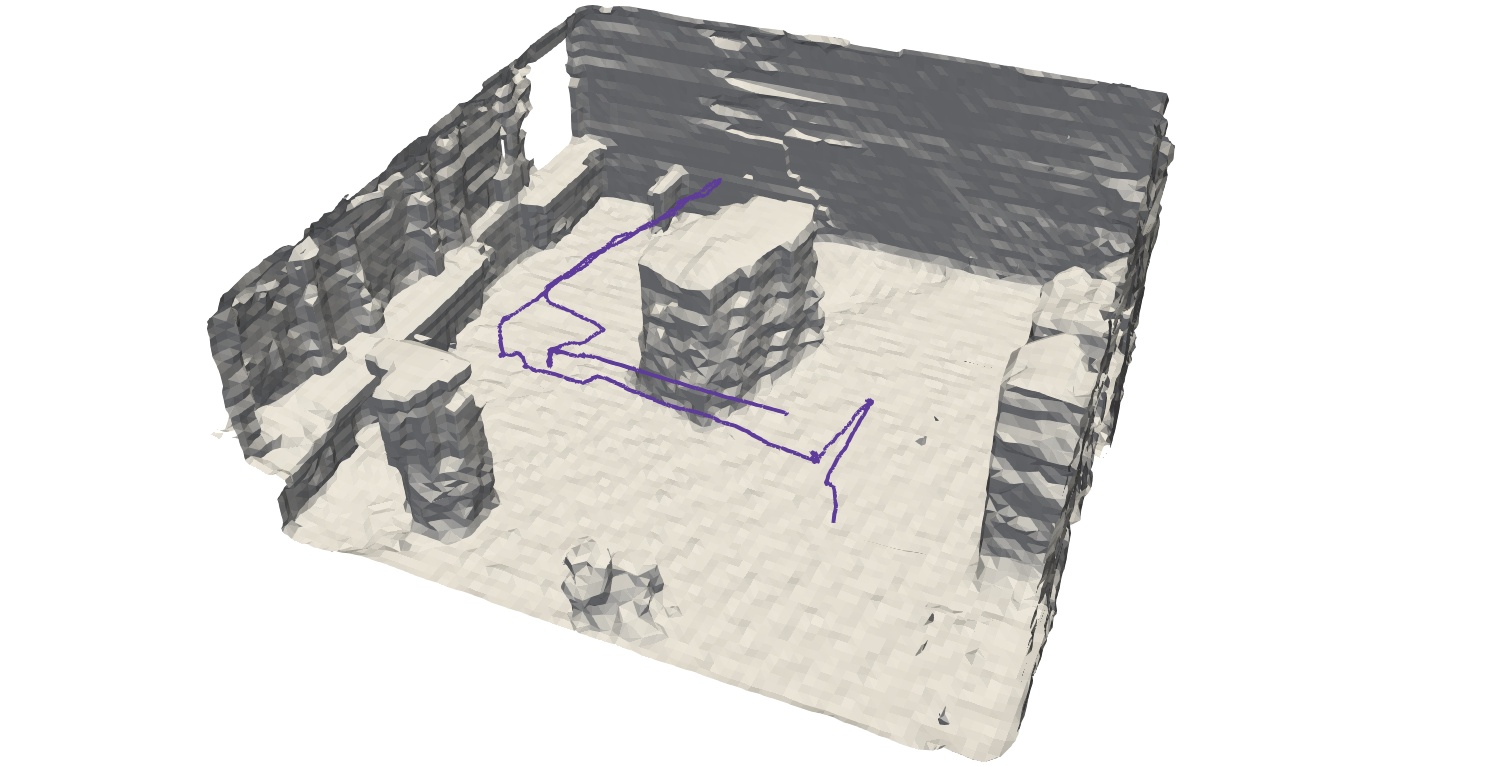}
    \includegraphics[width=0.493\columnwidth,trim={7.5cm, 0, 11.1cm, 0},clip]{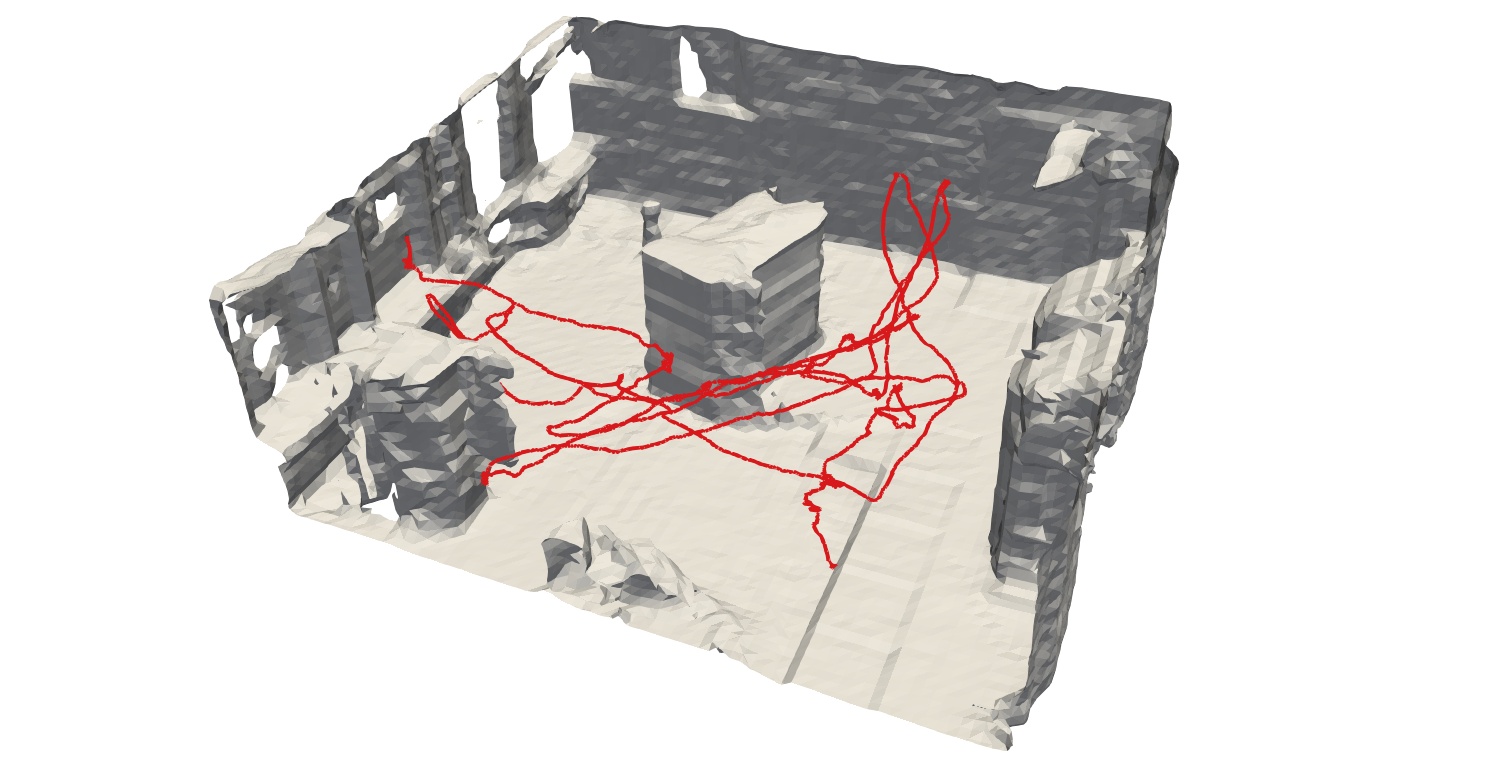}
    \caption{MAV trajectory and live-reconstructed mesh at 10 cm resolution for the real-world experiments using a LiDAR (left) and a depth camera (right).}
    \label{fig:rw-reconstruction}
\end{figure}

Keeping track of frontiers, the boundaries between observed and unobserved space, becomes challenging when using submaps, as a large number of them may be involved.
Recent works have proposed hierarchical exploration methods, distinguishing between local and global exploration planning~\cite{schmid2021unified,dang2019explore}.
In our work, we show that a more complex hierarchical approach is not necessary for fast and efficient exploration.

We propose the following contributions:
\begin{itemize}
    \item A fast and lightweight 3D exploration framework leveraging SLAM and submapping to account for odometry drift using loop closures.
    \item An efficient global frontier update method, allowing for accurate and complete large-scale exploration, without having to distinguish between local and global planning.
    \item The proposed system seamlessly supports MAVs equipped with either LiDAR or depth camera sensors out-of-the-box.
    \item Quantitative and qualitative evaluation in simulation and in the real world, showing faster exploration, reduced resource usage, and more consistent and complete reconstructions compared to a state-of-the-art method. 
\end{itemize}

\section{Related Work}
\label{chapter:related_work}

\subsection{Multi-Sensor SLAM}
Performing autonomous exploration in environments without prior infrastructure requires estimating the robot's state.
VI-SLAM achieves this given camera images and measurements from an Inertial Measurement Unit (IMU), which are especially vital for MAVs, platforms capable of highly dynamic movements, which might cause purely visual approaches to fail due to losing track.
VI-SLAM systems are often classified into filter-based, e.g.~\cite{mourikis2007multi, geneva2020openvins}, and optimization-based approaches~\cite{OKVIS2,campos2021orb,usenko2019visual,qin2019general}.
The latter tend to yield higher accuracy and usually formulate pose estimation as a factor graph optimization problem, minimizing residuals of IMU integration errors and visual reprojection errors.
To reduce the drift of VI-SLAM systems caused by the integration of noisy IMU measurements, fusing LiDAR-based residuals has been extensively investigated.
Many works adopted the early geometric LiDAR-based feature residuals (edge and plane) formulated in LOAM~\cite{loam}, e.g.~\cite{shan2021lvi}. However, direct or Iterative Closest Point (ICP) -based variants have also proven effective~\cite{zheng2022fast,vizzo2023kiss}.
In our work, we use our LVI-SLAM system as presented in~\cite{boche2024tightlycoupled}.

\subsection{Volumetric Mapping}
An important design decision in autonomous exploration systems is the map representation used.
Dense 3D maps, usable for downstream tasks, can be obtained from volumetric mapping.
The pioneering work KinectFusion~\cite{newcombe2011kinectfusion} represents the map as a Truncated Signed Distance Field (TSDF).
Unfortunately, TSDF-based maps are unable to explicitly represent \free space, making them unsuitable for navigation tasks.
Using Euclidean Signed Distance Fields (ESDF) instead is one way to solve this issue, as e.g.\ in \textit{voxblox}~\cite{oleynikova2017voxblox}.
Another line of work uses occupancy mapping, representing the map as a discrete grid of occupancy probabilities.
A large number of works follow \textit{Octomap}~\cite{octomap} in using an octree data structure as the underlying map representation.
Several extensions, such as \textit{UFOMap}~\cite{duberg2020ufomap} and \seTwo~\cite{SE2}, have been proposed since then.
Both approaches explicitly store \unknown space and support adaptive-resolution; mapping the 3D space only up to the required detail.
In our work, occupancy submaps are created using \seTwo~\cite{SE2}.

\subsection{Autonomous Exploration}
Related works in robotic exploration are commonly categorized into sampling-based and frontier-based methods.

The concept of frontiers was introduced in~\cite{yamauchi1997frontier}.
During the incremental reconstruction of an environment, frontiers are defined as the boundaries between known \free and \unknown space and indicate areas that potentially lead to a large gain of information when observed.
This concept found adoption in a large number of subsequent works, e.g.~\cite{cieslewski2017rapid,gao2018improved}.

Sampling-based approaches on the other hand sample a number of candidate viewpoints or paths and select the next goal based on the computation of a utility function, following the pioneering work of~\cite{bircher2016receding}. Candidate next views are sampled in free space and the next best view is selected based on the unknown volume potentially observed from the candidate over the path length.
The efficiency of these methods can be improved by informed sampling, e.g.\ in the proximity of frontiers~\cite{dai2020fast,duberg2022ufoexplorer,zhong2021information}, or other utility functions~\cite{schmid2020efficient,batinovic2022shadowcasting}.

To deal with large-scale environments, recent approaches have proposed hierarchical or multi-stage designs that distinguish between local and global planning.
In~\cite{selin2019efficient}, local planning is handled by a Rapidly-exploring Random Tree (RRT) that determines the next best view.
The global planner is modelled as a Gaussian Process that stores the gain of non-executed viewpoints.
\textit{GBPlanner}~\cite{dang2019explore} also follows this two-stage pattern and consists of a local RRT*-based planner and a global graph-based planner.
The global graph guides the robot towards frontiers, when the local planner reaches a dead-end. Furthermore, \cite{dang2019explore} imposes additional safety constraints on the selected path.

Regardless of the design, accurate live pose estimates are required, typically obtained from SLAM.
Several approaches have been proposed to overcome the deterioration of the map quality due to drift.
\cite{zhang2022exploration} addresses this issue by computationally expensive de- and re-integration of measurements upon visual loop closures. Furthermore, exploration is actively steered towards potential loop closure candidates.
Another line of works applies submapping strategies.
The global map is represented as a collection of submaps which can be re-arranged upon loop closures or pose-graph optimisation, as presented in~\cite{boche2024tightlycoupled,voxgraph,cblox,oxfordSubmappingExtended}.

A successful combination of submapping and hierarchical exploration has been presented in \glocal~\cite{schmid2021unified}. Submaps based on \voxgraph~\cite{voxgraph} are periodically created and updated in a pose-graph optimization which allows handling loop closures.
In \glocal, the MAV first explores its surroundings using a local, RRT*-based planner.
Once the utility of nearby regions becomes low, the global planner guides the robot towards global frontiers. Traversability graphs from previously completed submaps allow for efficient global path planning.\looseness=-1

Just like \glocal, our system builds upon the concept of submaps.
In contrast to the aforementioned approaches, we do not need to make a distinction between local and global planning.
Enabled by efficient updates of the local and global frontiers, a unified global planner can be applied, favoring nearby candidate next poses through the utility function design.

\begin{figure*}[ht]
    \centering
    \includegraphics[width=\textwidth]{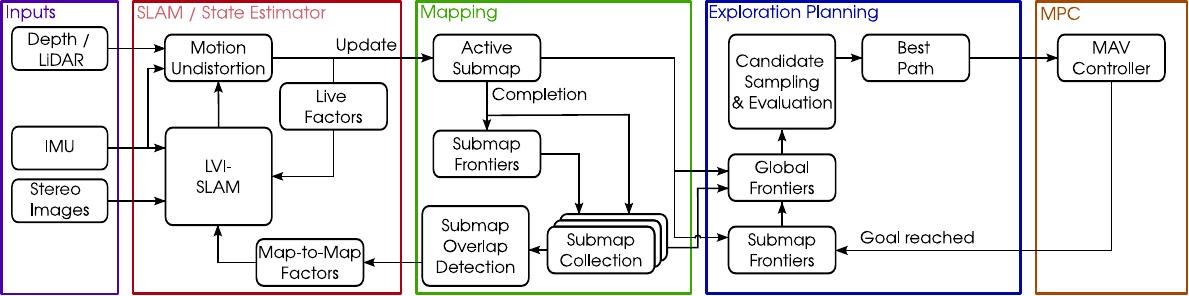}
    \caption{
    Overview of the system's main modules:
    (i) a VI-SLAM system, optionally with LiDAR, that estimates the MAV's state based on sensor inputs and submaps,
    (ii) an occupancy mapping backend dividing the global volume into spatially bounded submaps and keeping track of a submap collection,
    (iii) an exploration planning module computing the next best path based on global frontiers of all submaps and
    (iv) a linear MPC executing the planned paths.\looseness=-1
    }
    \label{fig:system-overview}
\end{figure*}

\section{Problem Statement}
\label{chapter:preliminaries}

In this work, we aim to build an accurate, complete and globally consistent volumetric representation of the environment using a fully autonomous MAV equipped with either a LiDAR or depth camera.
To enable scaling to large environments, we represent the environment with submaps.

\subsection{Environment Model}
The static environment is modeled as a bounded volume $V \subset \R^3$ where each point $\vect{v} \in V$ has an associated occupancy probability $P_o(\vect{v})$.
The occupancy of all points $\vect{v} \in V$ is initially \unknown, defined as $P_o(\vect{v})=0.5$.
Due to the environment's geometry as well as the MAV and sensor characteristics, there can be points $V_\mathrm{unob} \subset V$ that are unobservable.
Thus exploration only considers the observable part of the environment $V_\mathrm{obs} = V \setminus V_\mathrm{unob}$.
The goal of exploration is to create a collection of submaps $\submaps = \left\{ M_i \subseteq V, \, i \in \{1 \dots n_m\} \right\}$ so that $\bigcup\limits_{i=1}^{n_m} M_i = V_\mathrm{obs}$, while updating the occupancy probability of all $\vect{v} \in V_\mathrm{obs}$ to either \free or \occupied.\looseness=-1

\subsection{MAV Model}
The MAV state $\vect{x}$ consists of the position $\vect{r} \in V$, orientation $\vect{q} \in \SOthree$, and linear velocity $\vect{\upsilon} \in \R^3$.
For exploration and path planning we consider a portion of the MAV's full state, $\vecth{x} = [\vect{r}^T \psi]^T \in V \times [-\pi, \pi)$, where $\psi$ is its yaw angle with respect to the world frame $\cframe{W}$.
We also assume the MAV to be enclosed in a sphere of radius $R \in \R^+$ centered at $\vect{r}$, and to have a maximum linear velocity $\upsilon_{\max} \in \R^+$ and a maximum yaw rate $\omega_{\max} \in \R^+$.
The MAV is equipped with a LiDAR sensor or a depth camera with horizontal and vertical field of view $\alpha_h \in \left( 0, 2\pi \right]$ and $\alpha_v \in \left( 0, \pi \right]$ respectively, and a measurement range $\left[ d_{\min}, \, d_{\max} \right] \subset \R^+$.

\section{System Overview}
\label{chapter:system_overview}

Our system combines VI-SLAM, dense 3D occupancy mapping, exploration planning and an MAV Model Predictive Controller (MPC).
To deal with the accumulated VI-SLAM drift and achieve locally consistent and accurate maps, we employ spatially bounded submaps.
Using SLAM-estimated poses, LiDAR measurements or depth images are integrated into the currently active submap. Upon completion, submaps and their local frontiers are saved in a submap collection.
Global frontiers are only updated as needed, using the sumbap-local frontiers.
This allows using a single, global exploration planner, making it unnecessary to distinguish between local and global planning, as other methods do.
The global frontiers are used in a sampling-based method to determine the next best path from an information gain perspective, which is then tracked by the MPC.
Our method does not require a GPU, allowing its deployment on small and resource-constrained platforms.
In the following we describe the individual system modules, shown in \cref{fig:system-overview}.

\subsection{State Estimator}
We use OKVIS2~\cite{OKVIS2} as our state estimator, a state-of-the-art sparse, optimization-based VI-SLAM system. 
It receives stereo grayscale image pairs and inertial measurements and produces state estimates at IMU rate which include the MAV state $\vect{x}$ and IMU biases.
In the case of MAVs equipped with LiDARs, we also integrate our previous work~\cite{boche2024tightlycoupled}, which formulates residuals based on occupancy maps and their gradients to optimize for consistency between (a) incoming measurements and previous submaps, and (b) overlapping submaps.
The reader is referred to~\cite{boche2024tightlycoupled} for details.

\subsection{Occupancy Mapping}
We use \seTwo~\cite{SE2} for volumetric occupancy maps using octrees.
\SeTwo explicitly represents \free space and propagates minimum and maximum occupancy values to the octree root, allowing safe and efficient path planning.
The original \seTwo only allows integrating depth images, either from a depth camera or by projecting a structured LiDAR scan.
In order to support dynamically moving LiDAR sensors, we adopt the integration scheme from~\cite{boche2024tightlycoupled}.

\SeTwo forms the basis of our submapping framework, as in~\cite{boche2024tightlycoupled} and~\cite{digiforest}.
New submaps are generated based on the geometric overlap in the case of LiDAR sensors, as in~\cite{boche2024tightlycoupled}, or based on visual keyframes in the case of depth cameras, as in~\cite{digiforest}.
In both cases submaps are anchored to OKVIS2 keyframe states and are transformed along with them on pose optimization, including loop closures.
Once a new submap is created, the previous one is frozen, allowing only rigid transformations for the remainder of the mission.

\subsection{Exploration Planning}
We extend the exploration planner for monolithic maps we proposed in~\cite{dai2020fast} to account for submaps.
The key ideas of this planner remain unchanged and are re-stated here for convenience.
Candidate next positions $\vecth{r}_j \in V, \, j \in \left\{1 \dots n_c\right\}, \, n_c \in \N^+$ are sampled close to frontiers, since they correspond to regions that will expand the map if observed.
A path is planned from the current MAV position to each candidate and its estimated duration $t_j$ is computed assuming the MAV flies at $\upsilon_{\max}$ and $\omega_{\max}$.
A $360^{\circ}$ map entropy raycast is performed from each candidate position $\vecth{r}_j$ producing a gain image.
Given the sensor's horizontal field of view $\alpha_h$, a sliding window optimization is performed on the gain image to determine the yaw angle $\psi_j$ resulting in the highest potential gain $g_j$.
The utility of each candidate $j$ is computed as $u_j = g_j\,/\,t_j$, essentially maximizing information gain over time.
The candidate with the highest utility becomes the next goal and the process is repeated once the MAV reaches it.

The exploration planner used in this work differs from~\cite{dai2020fast} in two aspects.
First, we compute global frontiers from the collection of submaps $\submaps$, presented in \cref{sec:frontiers} as one of our core contributions.
Second, the yaw optimization over the gain image is performed differently for sensors with $\alpha_h = 360^{\circ}$, which~\cite{dai2020fast} is not designed to handle.
In this case the candidate yaw angle $\psi_j$ can be chosen arbitrarily while the candidate gain $g_j$ is just computed over the whole image.
For simplicity, we compute $\psi_j$ using a sliding window corresponding to a horizontal field of view smaller than $360^{\circ}$.

\subsection{MAV Controller}
\label{sec:mpc}
We use a linear MPC based on~\cite{tzoumanikas2019fully} for MAV trajectory following, with modifications as described in~\cite{digiforest}.
In short, the MPC is modified to ensure correct trajectory tracking even in the case of major odometry changes, such as loop closures.
This is achieved by anchoring trajectories to OKVIS2 keyframe states and elastically deforming them as the keyframe states are updated over time.

\section{Global Frontier Computation}
\label{sec:frontiers}

The exploration planner samples candidate next positions close to frontiers, thus, a set of global frontiers, considering all submaps and their overlaps, must be computed.
Frontiers in one submap corresponding to fully observed regions in another submap are not global frontiers.

To achieve efficient global frontier computation, we (i) keep track of submap-local and global frontiers separately, and (ii) update both only as needed.
The local frontiers of the currently active submap are updated before each global frontier computation.
Local frontiers are also updated upon submap completion and then remain frozen for the remainder of the mission.
In both cases, only the submap regions modified since the last local frontier computation are considered.
Global frontiers are re-computed at each exploration planner iteration, considering all currently known local frontiers.
Since global frontiers are only used for sampling candidate next views, this scheme ensures no unnecessary computations are performed.
The following describes the local and global frontier computation in detail.

\subsection{Local Frontier Computation}
For each submap $M_i$ we keep track of a set of local frontiers $\frontiers_i^k$ at timestep $k$, by considering $M_i$ in isolation from other submaps.
Local frontiers are \free voxels that (i) have at least one \unknown face neighbor voxel or (ii) are at the submap boundary.
The latter case is essential for correct global frontiers when using bounded-extent submaps.

Only the local frontiers $\frontiers_i^k$ of the currently active submap are updated. This update is done in two steps.
First, all preexisting local frontiers $\frontiers_i^{k-1}$ are re-tested so that the out-of-date frontiers $\frontiers_{i,\mathrm{stale}}^{k-1}$ are detected.
Then, the set of new frontiers $\frontiers_{i,\mathrm{new}}^k$ is computed by
only testing voxels that were modified since $\frontiers_i^{k-1}$ was computed.
Finally, the local frontiers at timestep $k$ are computed as
\begin{equation}
    \frontiers_i^{k} = \bigl( \frontiers_i^{k-1} \setminus \frontiers_{i,\mathrm{stale}}^{k-1} \bigr) \cup \frontiers_{i,\mathrm{new}}^k.
\end{equation}
The detection of stale frontiers $\frontiers_{i,\mathrm{stale}}^{k-1}$ can be performed in parallel with the computation of the new frontiers $\frontiers_{i,\mathrm{new}}^k$.

\subsection{Global Frontier Computation}
The global frontiers $\frontiers_g^k$ at timestep $k$ are computed as
\begin{equation}
    \frontiers_g^k = \bigcup\limits_{i=1}^{n_m} \frontiers_{g,i}^k,
\end{equation}
where $\frontiers_{g,i}^k \subseteq \frontiers_i$ are the local frontiers of $M_i$ that are also global frontiers at timestep $k$, and $\frontiers_i$ are the last-known frontiers of $M_i$.
$\frontiers_{g,i}^k$ is computed by testing each frontier in $\frontiers_i$ against all other submaps
\begin{equation}
    \frontiers_{g,i}^k = \Bigl\{ f_i \in \frontiers_i, \ \bigwedge\limits_{l\ne i} h(f_i, M_l) \Bigr\},
\end{equation}
where the Boolean-valued function $h(f_i, M_l)$ indicates whether frontier $f_i$ of $M_i$ is also a frontier when considering $M_l$.
$h(f_i, M_l)$ is \true iff $f_i \notin M_l$ or $f_i$ corresponds to an \unknown region of $M_l$ or $f_i$ corresponds to a local frontier of $M_l$.
The computation of $\frontiers_{g,i}^k$ can be performed in parallel for each submap.

\begin{figure}[t]
    \centering
    \includegraphics[width=\columnwidth]{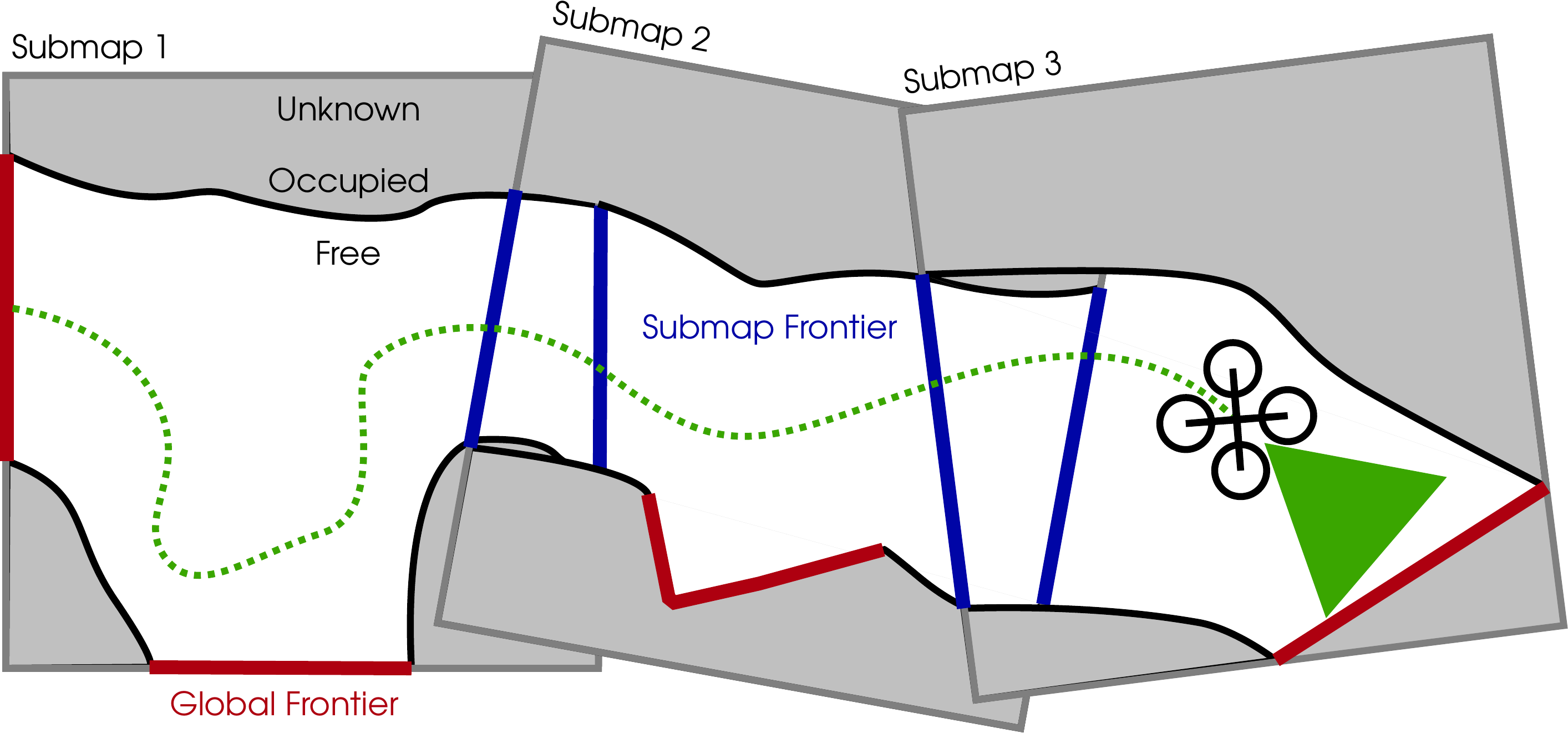}
    \caption{Local-only (blue) and global (red) frontiers in a set of submaps. The local-only submap frontiers correspond to regions mapped in other submaps and are thus not considered to be global frontiers.}
    \label{fig:frontiers}
\end{figure}

\cref{fig:frontiers} shows an example of local and global frontiers.

\section{Evaluation}
\label{chapter:results}

\subsection{Simulation Results}
\label{sec:results_sim}

\begin{figure}[htb]
    \centering
    \includegraphics[width=\columnwidth]{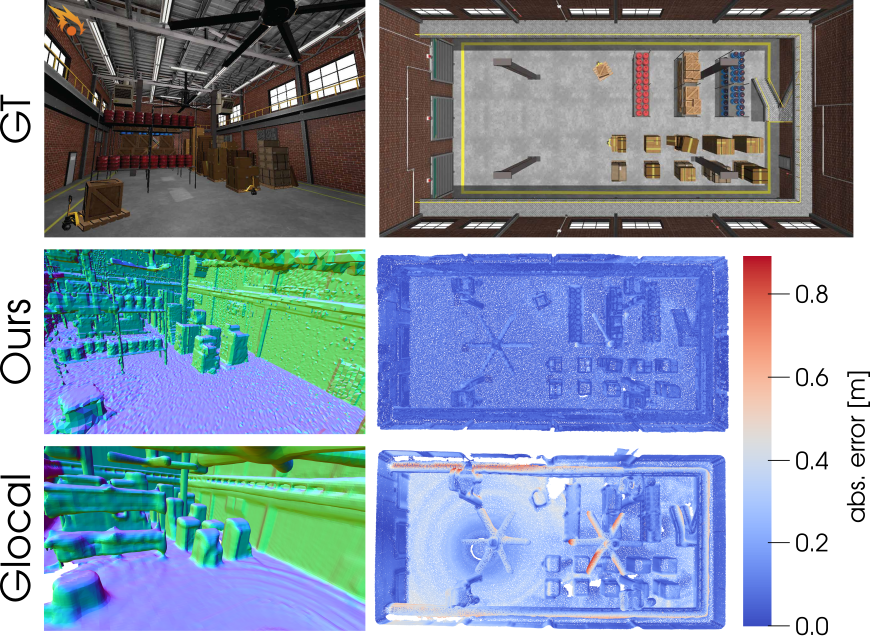}
    \caption{Perspective and top-down visualisation of the \textit{Depot} environment.
    Top: Ground-truth.
    Middle and Bottom: Final reconstruction (left) and mesh accuracy (right), using our approach and \glocal respectively.
    }
    \label{fig:sim-environments}
\end{figure}

Our simulation setup consists of Gazebo~\cite{gazebo} and the PX4 autopilot~\cite{px4} simulated in software, mimicking the setup of the real-world MAV.
The simulated MAV is based on the RMF-Owl~\cite{de2022rmf}, a $0.38\times0.38\times0.24$ m quadcopter equipped with a LiDAR sensor and an Intel RealSense D455 RGB-D camera.
The evaluation is done in
the $30 \times 15 \times 9$ m warehouse-like environment \textit{Depot}, shown in \cref{fig:sim-environments}.
All simulated experiments were conducted on a computer with an Intel Core i7-1165G7 CPU and 32 GB of RAM.

We compare our method with the state-of-the-art, submap-based, 3D exploration planner \glocal~\cite{schmid2021unified}, in terms of exploration speed, environment reconstruction accuracy and completeness, as well as CPU and memory usage.
\glocal requires a LiDAR sensor and performs submap alignment using \voxgraph~\cite{voxgraph}, similarly to our approach based on~\cite{boche2024tightlycoupled}.
Thus, we only evaluate LiDAR-based exploration in simulation.
Both approaches receive poses from OKVIS2, including visual loop closures.
Our approach includes submap alignment based on~\cite{boche2024tightlycoupled}, whereas \glocal uses map-to-map alignment from~\cite{voxgraph}.
We use the MAV controller described in \cref{sec:mpc} for both pipelines.
We made an effort to use the same parameters for both methods where applicable, the most important of which are listed in \cref{tab:sim_params}.

\begin{table}[htb]
    \centering
    \begin{tabular}{ll|ll}
    Map resolution & 0.1 m &
    LiDAR resolution & $360 \times 180$ \\
    $R$ & 0.5 m &
    $\alpha_h$, $\alpha_v$ & $360^\circ$, $90^\circ$ \\
    $\upsilon_{\max}$ & 0.5 m/s &
    $d_{\min}$, $d_{\max}$ & 1 m, 10 m \\
    $\omega_{\max}$ & 0.5 rad/s & $n_c$ (n/a to \glocal) & 20
    \end{tabular}
    \caption{Simulation experiment parameters.}
    \label{tab:sim_params}
\end{table}

\subsubsection{Explored Volume}
\begin{figure}[htb]
    \centering
    \includegraphics[width=\columnwidth]{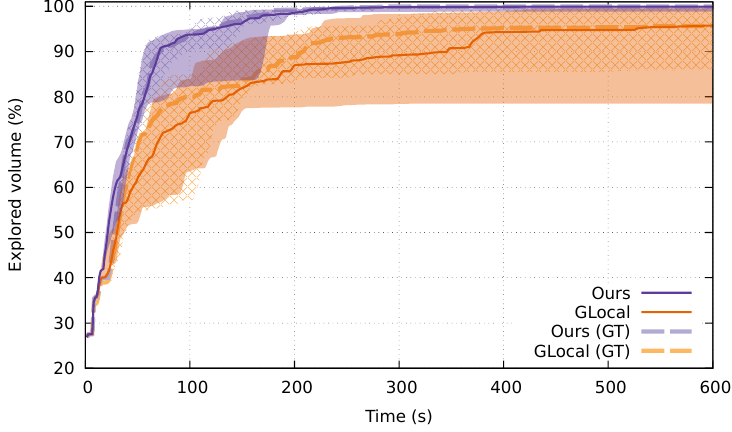}
    \caption{Explored volume median, $10^{th}$ and $90^{th}$ percentiles for 10 runs in the \textit{Depot} environment. Solid for SLAM, dashed/hatched for ground-truth runs.\looseness=-1}
    \label{fig:volume_depot}
\end{figure}

In order to avoid discrepancies in the explored volume due to different mapping frameworks, we record the LiDAR measurements and corresponding ground-truth MAV poses. We use them to construct a monolithic \seTwo map, measuring the explored volume after each integration.
For each method, we performed 10 runs using poses from OKVIS2 and another 10 runs using ground-truth poses as a baseline.
The median, $10^{th}$ and $90^{th}$ percentiles of the percentage of the total volume explored by the two methods is shown in \cref{fig:volume_depot}.
Our method outperforms \glocal in terms of exploration speed, consistency and final environment coverage.
While our approach nearly always explores $100\%$ of the volume within 300 s, \glocal sometimes fails to explore the whole volume.
Using ground-truth poses results in less variance between runs, which can be attributed to the more consistent maps resulting in fewer frontiers and more free space for the MAV to navigate through.

\subsubsection{Safety Evaluation}
\begin{figure}[t]
    \centering
    \includegraphics[width=\columnwidth]{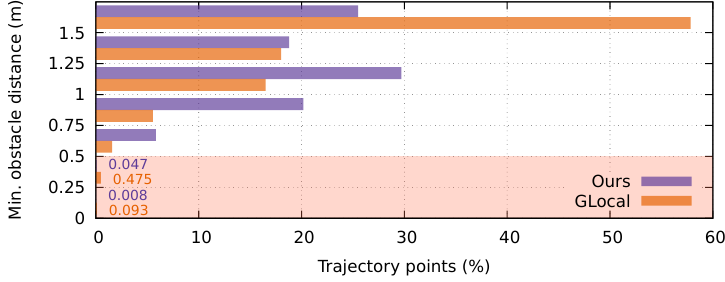}
    \caption{Histogram of minimum distance of the MAV to obstacles across 10 simulated missions with SLAM poses and a safety radius of 0.5 m. The bin size is 0.25 m.}
    \label{fig:collision_plot}
\end{figure}

We evaluate the safety of the exploration planner by computing the minimum distance between the \textit{Depot} ground-truth mesh and every point of the ground-truth MAV trajectory.
\cref{fig:collision_plot} shows a histogram of these distances across all 10 missions.
Our approach in general succeeds to navigate safely even in proximity to obstacles while \glocal violates the safety radius $R=0.5$ m significantly more often, leading to a higher risk of collisions.

\subsubsection{Reconstruction Quality}
We also evaluate the quality of the final mesh reconstructions at a 0.1 m resolution obtained from a post-processing step.
In our method, the final map reconstruction is computed using poses estimated after a final bundle adjustment to provide the best possible accuracy. 
We use the ground-truth \textit{Depot} mesh to compute the root-mean-square-error (RMSE) of the reconstruction.
The completeness is computed as the percentage of the reconstruction within 0.2 m or 0.4 m of the ground-truth.
The mean RMSE and completeness of all 10 runs are presented in \cref{tab:reconstruction}. Our approach yields significantly better results in both metrics, showcasing it is possible to achieve safe path planning without sacrificing reconstruction quality. 
This is explained to some extent by the fact that the \textit{voxblox} mapping framework used by \glocal artificially inflates obstacles for safer path planning.
A visual comparison of the final reconstructions as well as their accuracy is presented in \cref{fig:sim-environments}.

\begin{table}[htb]
  \centering
  \begin{tabular}{lccc}
  & \multirow{2}{*}{RMSE (m) $\downarrow$} & \multicolumn{2}{c}{Completeness (\%) $\uparrow$} \\
  & & within 0.2 m & within 0.4 m \\
  \hline
  Ours & \textbf{0.112} & \textbf{86.07} & \textbf{97.73} \\
  GLocal & 0.219 & 44.19 & 76.97 \\
  \hline
  \dimt{Ours (GT)} & \dimt{0.095} & \dimt{89.89} & \dimt{97.98} \\
  \dimt{GLocal (GT)} & \dimt{0.213} & \dimt{45.29} & \dimt{78.20} \\
  \end{tabular}
  \caption{Mesh reconstruction RMSE and completeness using both SLAM and ground-truth (GT) poses in the Depot environment.}
  \label{tab:reconstruction}
\end{table}

\subsubsection{Resource Usage}

We also compare the computational resources required by our method and \glocal in \cref{tab:time_memory}.
Even though both methods compute frontiers only as needed, ours benefits from the more efficient map representation of \seTwo.
While we only compute the utility for a limited number of candidate next positions, in \glocal it is computed for each vertex of the tree used for local planning, requiring significantly more time.
The planning time includes utility computation, exploration planning, and path planning.
The high planning time in \glocal is due to a dedicated thread re-planning at a high rate, accumulating a large amount of CPU time, even after ignoring trivial planning iterations requiring less than 200 ms.
Finally, the smaller memory usage of our system is due to the efficiency of \seTwo maps.

\begin{table}[htb]
    \centering
    \begin{tabular}{lcccc}
    & Frontiers (s) & Utility (s) & Planning (s) & Memory (GB) \\
    \hline
    Ours & \textbf{16 $\pm$ 7} & \textbf{11 $\pm$ 6} & \textbf{134 $\pm$ 45} & \textbf{6.7 $\pm$ 1.0} \\
    \glocal & 52 $\pm$ 7 & 275 $\pm$ 84 & 421 $\pm$ 76 & 23.5 $\pm$ 1.8
    \end{tabular}
    \caption{Mean and standard deviation of per-mission frontier, utility, and path and exploration planning computation time and memory usage.}
    \label{tab:time_memory}
\end{table}

\subsection{Real World Experiments}

Real-world experiments were conducted to further showcase the effectiveness of the proposed exploration approach on resource-constrained platforms and demonstrate its applicability to both LiDAR sensors and depth cameras.
The experiments were carried out in an $8.0\times8.1\times3.7$ m $\approx 240$ m$^{3}$ room containing some tall obstacles.

\label{sec:results_real}
\begin{figure}[htb]
    \centering
    \includegraphics[width=0.4\columnwidth]{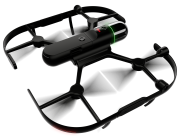}
    \hspace{1em}
    \includegraphics[width=0.4\columnwidth]{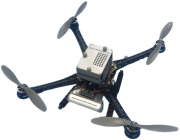}
    \caption{MAVs used for real-world experiments. Left: Leica BLK2Fly with LiDAR. Right: Custom MAV with Intel Realsense D455 RGB-D camera.}
    \label{fig:drones}
\end{figure}

For LiDAR exploration, we use a Leica BLK2Fly MAV, equipped with 5 cameras, 5 IMUs and a single-beam, dual-axis spinning LiDAR sensor.
For VI-SLAM, only the front, left, and bottom cameras and the bottom IMU are used.
The LiDAR has an effective frequency of 5 Hz and a $360^{\circ}$ field of view in both axes, although the MAV body occludes part of it, resulting in $\alpha_h$ and $\alpha_v$ both being less than $360^{\circ}$.
On-board filtering of the LiDAR point clouds reduces the number of points to $\approx$ 100,000 to 200,000 points per second.
As on-board resources are not made available for general use by the manufacturer, the sensor data is streamed via WiFi to a laptop where LVI-SLAM, mapping and exploration planning are running.
The laptop used for off-board processing has an Intel Core i7-13850HX CPU and 32 GB of RAM.

For depth camera exploration we use a custom-built quadcopter based on the Holybro S500 frame and equipped with an Intel RealSense D455 stereo RGB-D camera and an NVIDIA Jetson Orin NX 16 GB on-board computer.
All computations, including VI-SLAM, dense occupancy mapping, exploration planning and MPC are performed on-board.

Both MAV platforms are shown in \cref{fig:drones}.
The parameters used for the experiments are listed in \cref{tab:real_params}.

\begin{table}[htb]
    \centering
    \begin{tabular}{ll|l}
    & LiDAR & Depth camera \\
    \hline
    Map resolution & 0.1 m & 0.1m \\
    $d_{\min}$, $d_{\max}$ & 0.25 m, 10 m & 0.2 m, 4 m \\
    $R$ & 0.6 m & 0.6 m \\
    $\upsilon_{\max}$ & 1.0 m/s & 0.5 m/s \\
    $\omega_{\max}$ & - & 0.785 rad/s \\
    \end{tabular}
    \caption{Real-world experiment parameters.}
    \label{tab:real_params}
\end{table}

\begin{figure}[htb]
    \centering
    \includegraphics[width=\columnwidth]{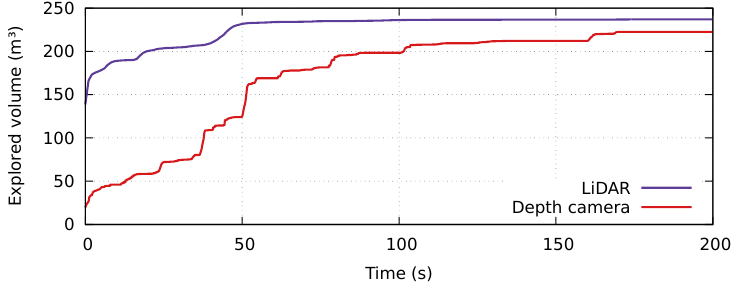}
    \caption{Explored volume over time for the two real-world experiments.}
    \label{fig:rw-volume}
\end{figure}

We conducted one experiment using each MAV platform.
The final mesh reconstructions and estimated MAV trajectories are shown in \cref{fig:rw-reconstruction} while \cref{fig:rw-volume} shows the observed volume over time.
As expected, due to its much larger field of view, the LiDAR-equipped MAV achieves a higher exploration speed and reaches an almost complete environment coverage within less than a minute.
Nevertheless, even using a depth camera with a much more limited field of view, we can almost fully explore the room in an efficient manner.

\section{Conclusion}
\label{chapter:conclusion}
In this work we propose an efficient and lightweight submap-based autonomous exploration method that we demonstrate to accept both depth cameras and LiDARs as input modalities.
Our method leverages submapping and VI-(LiDAR)-SLAM, in order to achieve accurate and consistent mapping despite odometry drift, thus ensuring safe operation.
The efficient computation of global frontiers from the aggregated submaps allows us to apply only one unified global planning approach, rendering the distinction between local and global planning that state-of-the-art methods employ unnecessary.
Compared to a state-of-the-art submap-based large-scale exploration framework, our method achieves faster exploration and more accurate environment reconstructions while being even more resource efficient.
It is further deployed on two real-world MAV platforms, one using a depth camera, and the other a LiDAR.\looseness=-1

In future work, we would like to integrate semantic information and Vision-Language Models (VLMs) to produce environment maps richer in information and more useful for downstream tasks.
We would also like to investigate integrating trajectory planning taking the MAV dynamics into account as well as safe MAV control strategies.

\bibliographystyle{IEEEtran}


\end{document}